\PassOptionsToPackage{unicode}{hyperref}
\PassOptionsToPackage{hyphens}{url}
\documentclass[
]{article}
\usepackage{xcolor}
\usepackage{amsmath,amssymb}
\usepackage[T1]{fontenc}
\usepackage[utf8]{inputenc}
\usepackage{textcomp}
\usepackage{lmodern}
\IfFileExists{upquote.sty}{\usepackage{upquote}}{}
\IfFileExists{microtype.sty}{% use microtype if available
  \usepackage[]{microtype}
  \UseMicrotypeSet[protrusion]{basicmath} % disable protrusion for tt fonts
}{}
\makeatletter
\@ifundefined{KOMAClassName}{% if non-KOMA class
  \IfFileExists{parskip.sty}{%
    \usepackage{parskip}
  }{% else
    \setlength{\parindent}{0pt}
    \setlength{\parskip}{6pt plus 2pt minus 1pt}}
}{% if KOMA class
  \KOMAoptions{parskip=half}}
\makeatother
\usepackage{longtable,booktabs,array}
\usepackage{multirow}
\usepackage{wrapfig}
\usepackage{changepage}
\usepackage{adjustbox}
\usepackage{makecell}
\usepackage{titling}
\usepackage{times}
\usepackage[top=1in, bottom=1in, left=1.5in, right=1.5in]{geometry}
\usepackage{calc} % for calculating minipage widths
\usepackage{etoolbox}
\makeatletter
\patchcmd\longtable{\par}{\if@noskipsec\mbox{}\fi\par}{}{}
\makeatother
\IfFileExists{footnotehyper.sty}{\usepackage{footnotehyper}}{\usepackage{footnote}}
\makesavenoteenv{longtable}
\usepackage{graphicx}
\makeatletter
\newsavebox\pandoc@box
\newcommand*\pandocbounded[1]{% scales image to fit in text height/width
  \sbox\pandoc@box{#1}%
  \Gscale@div\@tempa{\textheight}{\dimexpr\ht\pandoc@box+\dp\pandoc@box\relax}%
  \Gscale@div\@tempb{\linewidth}{\wd\pandoc@box}%
  \ifdim\@tempb\p@<\@tempa\p@\let\@tempa\@tempb\fi% select the smaller of both
  \ifdim\@tempa\p@<\p@\scalebox{\@tempa}{\usebox\pandoc@box}%
  \else\usebox{\pandoc@box}%
  \fi%
}
\def\fps@figure{htbp}
\makeatother
\usepackage{bookmark}
\IfFileExists{xurl.sty}{\usepackage{xurl}}{} % add URL line breaks if available
\usepackage{hyperref}
\hypersetup{
  hidelinks,
  pdfcreator={LaTeX via pandoc}}

\newcommand{\papertitle}{\textbf{Generative AI for Strategic Plan Development}}
\newcommand{\paperauthors}{\textbf{Jesse Ponnock}}
\newcommand{\paperaffiliations}{
Johns Hopkins University\\
\texttt{jponnoc1@jh.edu}\\
\emph{Originally prepared August 2023; submitted to arXiv August 2025}
}

\begin{document}

%\vspace*{2em}
\begin{center}
\rule{\textwidth}{4pt} \\[1.25em]  % Thicker top line and more space
\LARGE \textbf{\papertitle} \\[0.25em]
\rule{\textwidth}{1pt} \\[1.25em]
\large \textbf{\paperauthors} \\[0.5em]
\small \paperaffiliations
\end{center}

\vspace{2em}

\begin{center}
\Large \textbf{Abstract}
\end{center}

\begin{adjustwidth}{3.5em}{3.5em}
Given recent breakthroughs in Generative Artificial Intelligence (GAI)
and Large Language Models (LLMs), more and more professional services
are being augmented through Artificial Intelligence (AI), which once
seemed impossible to automate. This paper presents a modular model for
leveraging GAI in developing strategic plans for large scale government
organizations and evaluates leading machine learning techniques in their
application towards one of the identified modules. Specifically, the
performance of BERTopic and Non-negative Matrix Factorization (NMF) are
evaluated in their ability to use topic modeling to generate themes
representative of Vision Elements within a strategic plan. To accomplish
this, BERTopic and NMF models are trained using a large volume of
reports from the Government Accountability Office (GAO). The generated
topics from each model are then scored for similarity against the Vision
Elements of a published strategic plan and the results are compared. Our
results show that these techniques are capable of generating themes
similar to 100\% of the elements being evaluated against. Further, we
conclude that BERTopic performs best in this application with more than
half of its correlated topics achieving a ``medium'' or ``strong''
correlation. A capability of GAI-enabled strategic plan development
impacts a multi-billion dollar industry and assists the federal
government in overcoming regulatory requirements which are crucial to
the public good. Further work will focus on the operationalization of
the concept proven in this study as well as viability of the remaining
modules in the proposed model for GAI-generated strategic plans.
\end{adjustwidth}
\section{Introduction}\label{introduction}

\subsection{Background}\label{a.-background}

The purpose of this paper is to analyze the ability of GAI to develop
strategic plans for large scale government organizations. The proposed
framework decomposes the various modular tasks involved in developing a
strategic plan and evaluates the suitability of GAI to address these
needs. Further, industry-leading GAI techniques are used to model
outputs for a domain-specific use case and their performances are
evaluated to identify a leading solution.

Strategic planning is defined as ``a process in which an organization's
leaders define their vision for the future and identify their
organizations goals and objectives'' {[}1{]}. When executed effectively,
these goals and objectives will dictate the projects and activities that
the organization undertakes over the lifecycle of the plan. Because of
this relationship between visioning and business planning, a strategic
plan will deeply influence the direction of an organization and can be
solely responsible for the failure or success of achieving its mission.
In government, a strategic plan is a necessary tool in demonstrating
transparency in direction, accountability in elected officials, and good
stewardship of public funds. For this reason, in 1993 congress passed
the Government Performance and Results Act (GPRA) requiring all major
agencies to develop and maintain a strategic plan. In private industry,
strategic planning can be crucial in ensuring growth and profitability.
Notably, 71\% of fast-growing companies have strategic plans, business
plans, or similar long-range planning tools in place {[}2{]} and those
which do not consistently report weaker financial results than their
peers {[}3{]}.

Strategic planning is a multi-billion dollar industry led primarily by
large consulting firms due to the expertise required in developing a
strategy. It is estimated that the global strategy consulting market
will reach \$111.4 billion by 2031 {[}4{]}. The method for developing a
strategic plan can differ from case to case however a typical
implementation will involve a small group of professionals working for
several months, sometimes a year or more, at a project cost at times
accruing to millions of dollars. The steps taken by these individuals
include a series of discussions and analysis to extract key themes
around an organization's envisioned future and deriving a series of
goals and objectives from these findings. It is the hypothesis of this
research that with recent breakthroughs in fields such as natural
language processing, sentiment analysis, topic modeling, and GAI, a
system can be designed and developed to automate the generation of a
strategic plan with minimal manual human involvement, significantly
reducing the cost and time required to produce.

To begin exploring the potential for a GAI-generated strategic plan, we
must first understand the components required for such a product. The
below diagram depicts the components of a strategic plan, a brief
description of each, and an associated cognitive task behind the
creation of each.

\begin{figure}[htbp]
  \centering
  \includegraphics[width=\linewidth]{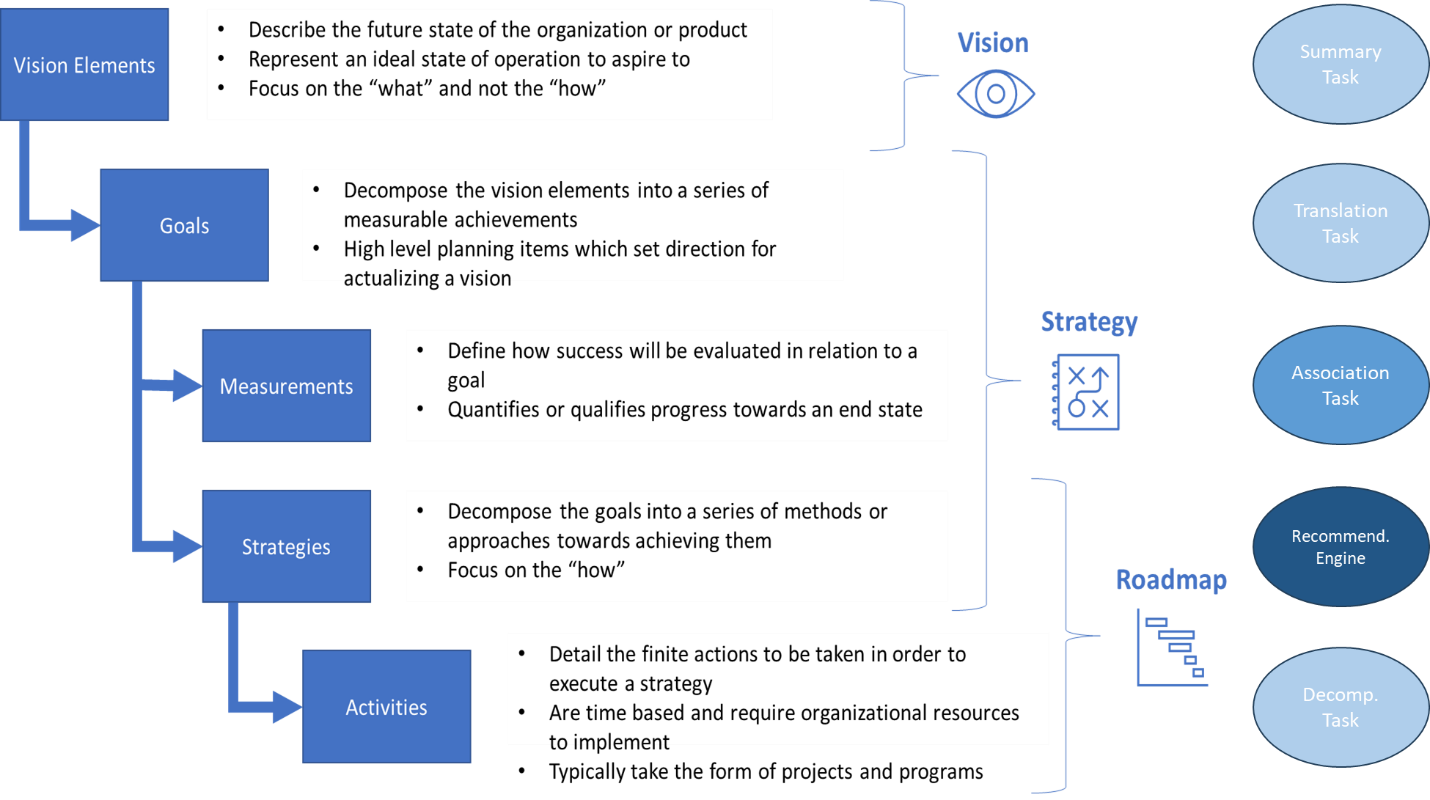}
  \caption{Cognitive Model for Strategic Plan Development}
  \label{fig:cognitive-model}
\end{figure}

The model above proposes five key components in developing a strategic
plan: Vision Elements, Goals, Measurements, Strategies, and Activities.
It indicates alternative groupings for these components to describe
their relationship to complementary artifacts which are common in the
field of strategy consulting. For example, Vision Elements are sometimes
compiled into a standalone document to describe an organization's Vision
or future state. Similarly, a series of strategies and activities will
often be compiled in an effort to develop a Roadmap. For the purposes of
this study, we will view all of these components in aggregate as the
strategic plan.

In the ovals that correspond to these 5 components to the right of the
diagram, we describe the proposed primary cognitive task associated with
developing each of these components. For example, development of the
Vision Elements, or future state, for an organization typically involves
stakeholder interviews with one or more senior executives and a review
of large collections of documentation from and about the organization in
an effort to identify future needs and potential modes of operation.
This task aims to collect large swaths of information which through
analysis is then summarized into key themes which will represent the
targeted future state of the organization. For example, if several
sources describe impacts of aging technologies, an analyst may identify
IT modernization as a theme for a Vision Element as part of the broader
Vision.

It should be noted that the contents of a strategic plan can differ by
implementation and is heavily driven by the design choices of its
implementor as well as the specific purpose the plan is looking to
serve. It is generally agreed upon however that at its core a strategic
plan will include a representation of a future state (referred to above
as the Vision), goals and corresponding measures, and some depiction of
actions or recommendations that will actualize the strategic direction
(referred to above as Strategies and Activities). The described model
above focuses on these core elements for the purposes of this study but
others may exist with differing terminology and scope within the areas
described.

The proposed primary cognitive tasks provide a structure to quantify the
largely qualitative problem of strategic plan development. It follows
that if we can break down the various tasks in this process into their
equivalent human cognitive requirements, and then solve for each of
those tasks using machine learning, we will have solved for the holistic
larger process. Viewing the results of this exercise tells us several
things about this exploration. First, it will not be a singular machine
learning model that will solve this problem holistically. Instead,
several different models will be required, addressing different tasks
within the larger problem, and some form of model orchestration will be
necessary to manage the models working in sequence. This is meaningful
but unsurprising, given this is often the case with solving large
problems using artificial intelligence, including recent breakthrough
products such as Chat-GPT {[}5{]}. Second, some tasks in this list
dictate more cognitive complexity than others and will in turn be more
challenging to address computationally. For example, a Translation Task
is considered a solved problem in AI, generally requiring little more
than an understanding of how to substitute words between two lexicons. A
Recommendation Task, however, is far more cognitively complex, often
requiring elements of situational awareness, trade-off analysis, and an
understanding of a potentially very large solution space and resulting
impact. An initial assessment of cognitive complexity is performed on
these tasks and is represented in the diagram by the shading of the
ovals, with the lighter color indicating less complexity and a darker
color indicating more.

Addressing every cognitive task presented in this diagram would be more
than a single study could account for. For this reason, this paper
focuses on the initial exploration into solving this larger problem:
presenting a model for how it would be addressed holistically and
proving the model's viability by evaluating an initial use case through
one of the cognitive tasks. In support of this initial exploration, we
evaluate the ability of state of the art NLP and GAI capabilities to
perform the Summary Task in support of identifying key themes for the
Vision within a strategic plan. This cognitive task was chosen based on
its low complexity and it being a dependency for many of the other tasks
downstream. Future studies will aim to explore the remaining cognitive
tasks represented in this diagram, providing an iterative approach to
solving the larger problem of GAI for strategic plan development.
Namely, this initial study will provide the following contributions:

\begin{enumerate}
\def\labelenumi{\arabic{enumi}.}
\item
  Cognitive model for strategic plan development, decomposing the core
  elements of a strategic plan and the primary cognitive functions
  required for their creation (see Figure 1, above)
\item
  Collected and curated dataset of text samples used for model training
  and analysis in developing the Vision for a strategic plan
\item
  Evaluation of methods in their application to the domain specific
  Summary Task of extracting information for Vision development
\end{enumerate}

\subsection{Literature Review}\label{b.-literature-review}

In relation to developing the Vision for a strategic plan, The United
Nations Strategic Planning Guide states that ``once the external and
internal inputs are analyzed, management sets the overall direction and
goal for the office or team'' {[}6{]}. External inputs consist of any
information from outside of an organization which provides an external
perspective on the challenges and trends that will impact the
organization. The data design for this study will look to mirror this
perspective by leveraging Government Accountability Office (GAO) reports
to represent this point of view. GAO reports consist of the findings
from a GAO study or audit, evaluating the efficiency, effectiveness, and
legality related to how an agency is operating. The described process of
analyzing this type of input involves identification of key trends or
themes within them which may qualify as desired characteristics of an
organization's future state. This is what is referred to as the Summary
Task in the diagram above. To replicate this process in this study we
evaluate state of the art approaches to topic modeling for its
effectiveness in identifying these key trends and themes and thus
generating these outputs for a strategic plan.

There is a significant body of work related to exploring the
applications and performance of topic modeling techniques. A 2020 study
investigates the performance of four prominent techniques in this area -
Latent Dirichlet Allocation (LDA), Non-negative Matrix Factorization
(NMF), Top2Vec, and BERTopic {[}7{]}. The study found several
characteristics of each technique which would impact their suitability
in application towards certain domains. LDA was found to favor text
which was geographically oriented, showing notable weighting towards
topics which reference countries or locations. When compared to NMF, LDA
was also found to produce more irrelevant topics whereas NMF produced
results more in line with human judgement. In a similar comparison,
BERTopic produced more specific and distinct topics whereas Top2Vec
showed overlapping concepts between topics. The study also indicates a
critical observation with Top2Vec in that, according the authors, it is
unqualified to work with small data sets, requiring by their estimation
a minimum of 1,000 documents to produce meaningful results.

In addition to application and performance, this body of research shows
evaluating topic modeling techniques to be a particularly challenging
area. One study explores this concept and concludes that results from
topic modeling techniques cannot be compared with a ``value'' and must
instead be qualitatively assessed by a human for meaningful
interpretation {[}8{]}. This is demonstrated in a recent 2023 study
where the author evaluates the quality of text generation from several
GAI models. The results are assessed and given a qualitative ranking
such as ``bad,'' ``good,'' ``mixed,'' or ``excellent.'' The author then
summarizes an overall assessment based on the aggregate quality of these
rankings and notes where improvements are found {[}9{]}.

Recent breakthroughs in use of LLMs for generative tasks, such as
Chat-GPT, call into the question the necessity of model training using
one of the techniques mentioned above. Chat-GPT is a LLM pre-trained to
follow an instruction and generate a content-based response. This
content can include a simple answer to a question, an executable segment
of computer code, the summarization of large bodies of text, and much
more. However, Generative Pre-Trained Transformer (GPT), the
foundational model upon which Chat-GPT operates, is shown to achieve
poor performance when attempting to generalize to out-of-distribution
tasks {[}10{]}. This means that Chat-GPT may be well suited for general
tasking but will struggle with tasks that are less common, such as those
specific to a domain, requiring niche language or expertise.

\section{Methods}\label{methods}

\subsection{Framework/Approach}\label{a.-frameworkapproach}

The purpose of this paper is to analyze the ability of GAI to develop a
strategic plan for large scale government organizations. In the
introduction, we present a cognitive model for developing a strategic
plan, breaking down its various elements and the cognitive tasks that
need to be replicated in order to achieve this. We converge on an
initial use case with a focus on simulating a Summary Task in support of
generating a Vision, or future state, the cornerstone of any strategic
plan. Through a targeted literature review we then make a case for using
topic modeling to generate the key themes for this Summary Task. Key
considerations related to data design and model selection are explored
which drive the approach taken for this study. The details of this
approach are explained further here, detailing methods employed to
derive the reported results.

\begin{figure}[htbp]
  \centering
  \includegraphics[width=\linewidth]{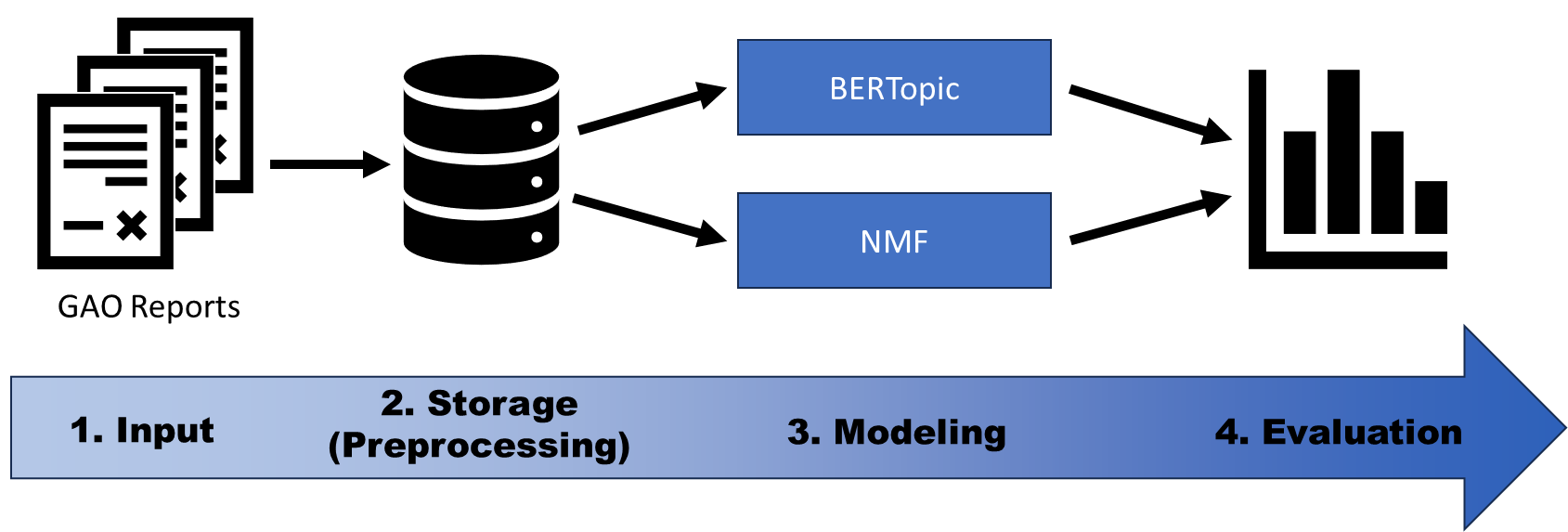}
  \caption{Theoretical Framework}
  \label{fig:theoretical-framework}
\end{figure}

This study uses GAO reports to represent perspectives on challenges and
trends facing an organization. GAO reports are available to download
from the GAO website, though large quantities are required as well as
reformatting for use with the identified modeling techniques, presenting
the need for specialized data collection and processing capabilities as
part of this design. Once collected and processed, this dataset is used
to train the selected models and outputs are generated and compared for
performance against a published strategic plan. For this evaluation, the
Department of Energy (DOE) Office of Legacy Management (LM) is selected
as a case study due to the availability of GAO reports pertaining to DOE
as well as the targeted focus of the organization. As such, the
following details the collection of DOE GAO reports, the generation of
topics pertaining to this organization, and the evaluation of results
against the most recent DOE LM Strategic Plan as of this study.

\subsection{Data Collection and Preprocessing}\label{b.-data-collection-and-preprocessing}

The GAO website hosts a publicly available library of GAO reporting over
the past 70+ years. However, these reports can only be accessed through
a search feature which displays a small number of results at a time,
with each result requiring further navigation (i.e., click-through) to
download the report. An automated means is necessary to download the
number of reports required for this study which this website does not
currently provide. To address this, we first evaluate the form of the
target URLs which allow for downloading the reports required. From
examining several examples, it becomes clear that the URL to download
each report is consistent except for the report's unique report number
appended to the end. This tells us that if we can collect a list of GAO
report numbers for the reports of interest, this URL convention can be
exploited through automation in order to obtain the reports in larger
numbers than the native search feature allows.

To collect this list of GAO report numbers, a web scraper is developed.
This web scraper leverages a Python library named Beautiful Soup which
is a tool capable of parsing HTML documents and extracting key items. In
this case, Beautiful Soup is pointed to the several pages of search
results on the GAO website and instructed to iterate through the HTML
elements that house the report numbers, saving them into a list.
Leveraging this list, a Python script is developed to take each report
number, append to the identified GAO website URL, and save the returned
file to a local folder.

As we learned through our literature review, BERTopic and NMF seem best
suited for the targeted application of this study however these
techniques require some meaningful considerations. BERTopic for example
performs clustering by document, with each cluster representing a topic.
This is less than ideal when attempting to create topics for long form
documents like a report which will likely include multiple topics of
interest. Similarly, like most machine learning models, both BERTopic
and NMF perform better with more instances of training data. To address
these considerations, we preprocess the data by dividing each report by
its pages, representing each page as a standalone document to be
clustered. Although effective in both addressing the potential for
multiple topics being in a report and increasing the number of training
samples, it should be noted that deciding where to separate the report
is a design choice which could impact the topic modeling results.
Dividing by page in this instance is a simple solution that meets the
needs of an initial evaluation of modeling techniques.

With these considerations accounted for, the final preprocessing step is
to convert these documents into a format which BERTopic and NMF can use
as inputs. This is necessary because the downloaded GAO reports are in
the form of PDF whereas both the techniques targeted accept a list of
strings as input. To address this, another Python script is created
using a library named PyPDF2 capable of splitting, merging, cropping,
and transforming the pages of PDF files. This script enables the
conversion of the downloaded reports into a list of strings, completing
the necessary preprocessing of data before modeling.

\subsection{Modeling}\label{c.-modeling}

We begin our modeling by examining the technique of Non-Negative Matrix
Factorization (NMF). The NMF process begins by calculating the Term
Frequency -- Inverse Document Frequency (TF-IDF) for our collection of
documents. TF-IDF is a statistical technique used to calculate the
importance of words within a corpus. Specifically, it calculates the
number of times a word appears in a document multiplied by the inverse
of the number of documents the term appears in throughout the collection
of texts. The assumption with TF-IDF is that a words importance is
inversely related to its frequency across a corpus. NMF decomposes a
matrix of calculated TF-IDF information (A) to a product of a
term-topics matrix (W) and a topics-documents matrix (H) {[}11{]}. For
these reasons, NMF is unique from other techniques used in this study in
that it is algebraic in its use of matrix factorization. Applying this
technique to our corpus produces the following topics.

\begin{figure}[htbp]
  \centering
  \includegraphics[width=\linewidth]{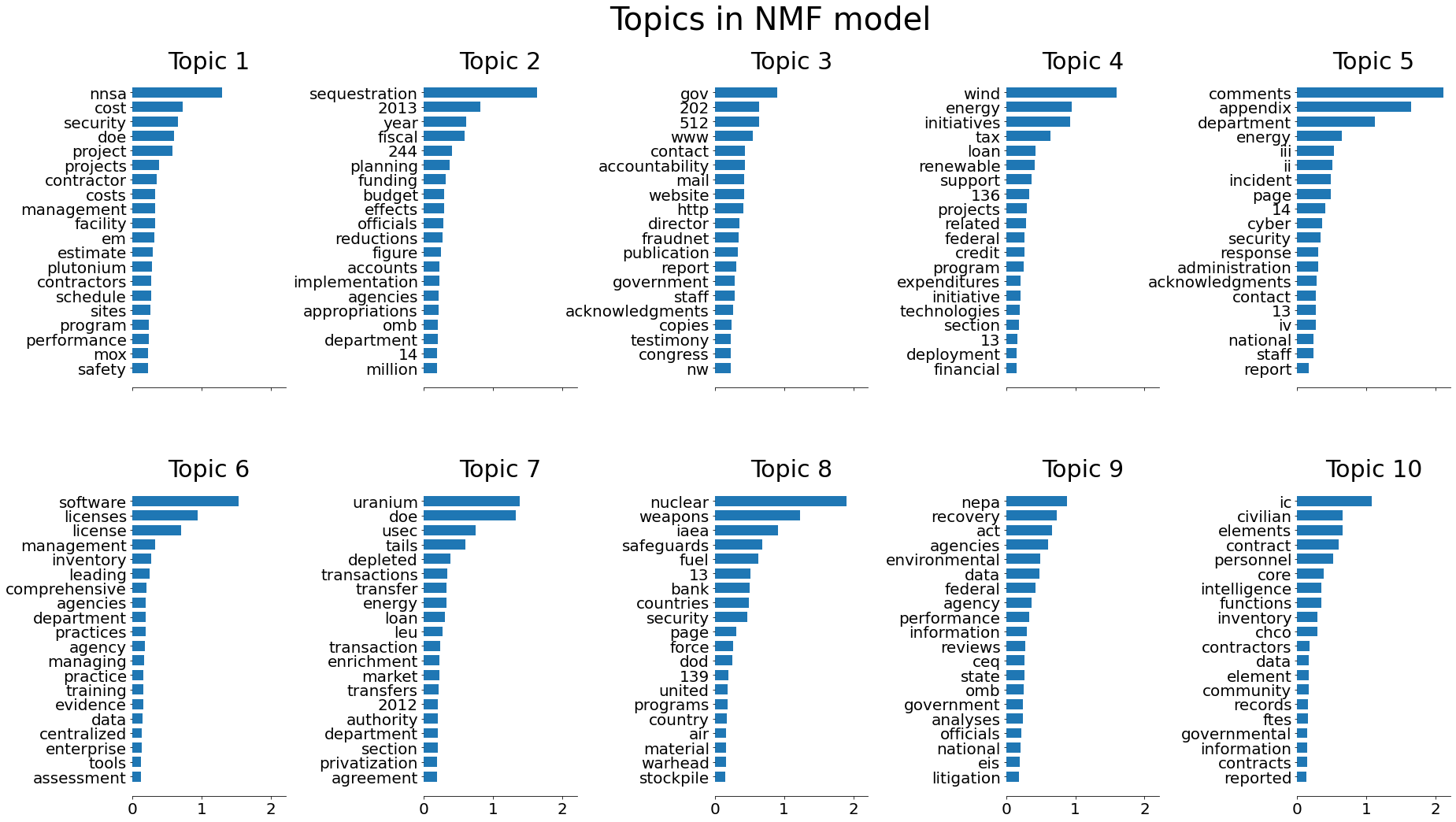}
  \caption{NMF Topic Representations of GAO Reports}
  \label{fig:nmf-topics}
\end{figure}

Figure 3 displays the top 10 topics resulting from application of the
NMF technique. Each topic presented displays its corresponding top words
resulting from the trained model along with a bar chart showing their
weights. These weights will become important to our analysis as it
demonstrates alternatives to the derived topic which have the potential
to be strengthened through fine-tuning. By examining this figure we see
that all topics have a clear leading candidate with the exception of
Topic 7 and Topic 9. We will analyze this output further and its
applicability to the domain of strategic plan development in the
following section.

BERTopic is another leading topic modeling technique that differs from
NMF in several ways. Most notably is its use of word embeddings based on
the Bidirectional Encoder Representations from Transform (BERT) language
model in generating word and sentence vector representations. Once the
corpus has been vectorized, BERTopic uses Uniform Manifold Approximation
and Projection (UMAP) to reduce the dimensionality of these embeddings
for optimized clustering with Hierarchical Density-Based Spatial
Clustering of Applications with Noise (HDBSCAN). With initial HDBSCAN
clusters generated, a custom variation of TF-IDF (the same technique
used in NMF) is deployed called Class - Term Frequency - Inverse
Document Frequency (c-TF-IDF) to generate topic representations. This
modified method works very similarly to TF-IDF except that the HDBSCAN
clusters are used to form classes and then inverse document frequency is
replaced by inverse class frequency to measure the importance of a term
to a class {[}12{]}. Training BERTopic against our corpus yields the
following topic representations.

\begin{figure}[htbp]
  \centering
  \includegraphics[width=\linewidth]{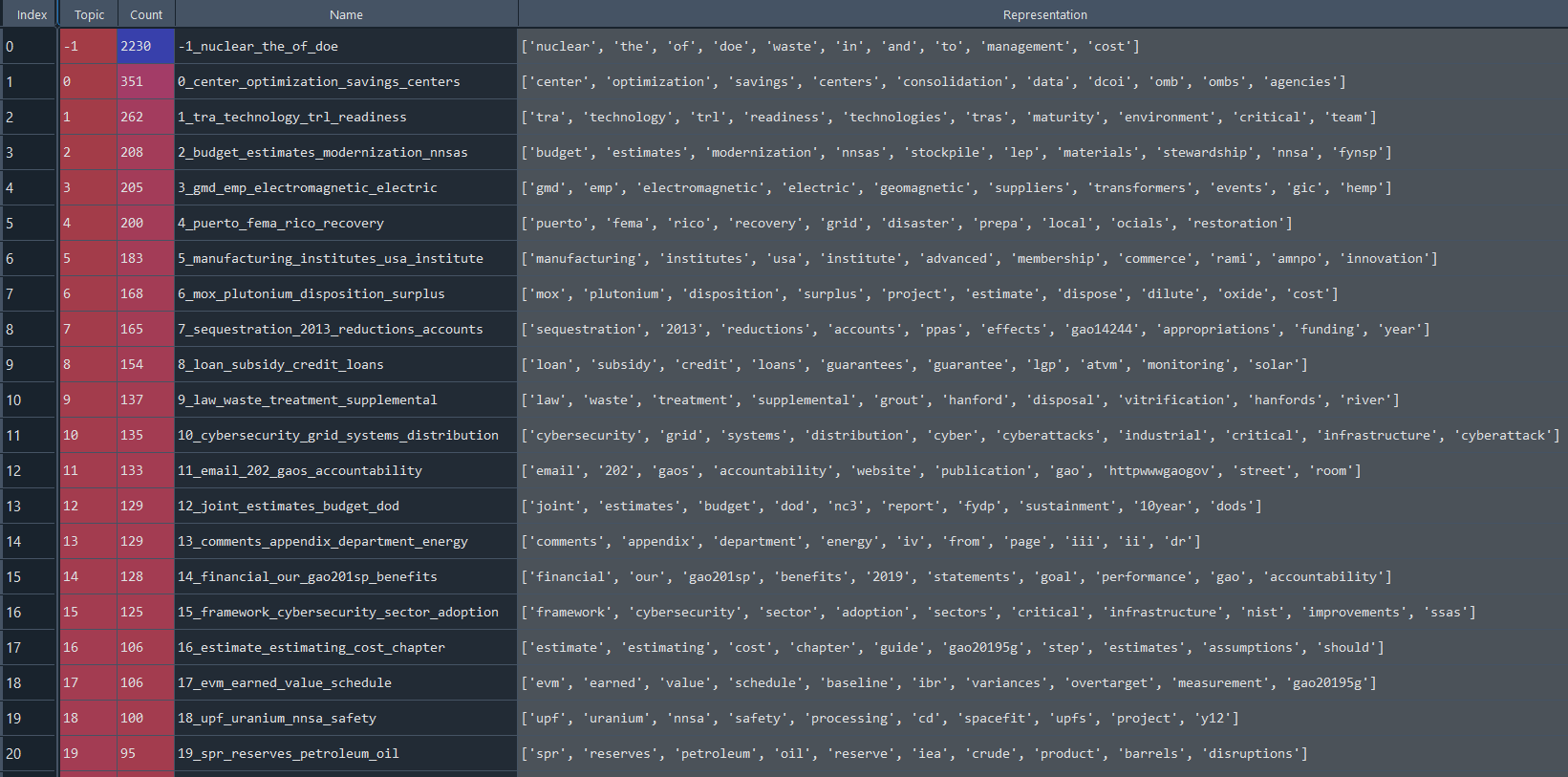}
  \caption{BERTopic Topic Representations of GAO Reports}
  \label{fig:bertopic-topics}
\end{figure}

The trained BERTopic model produces 347 topics in total and Figure 4 above
displays the top 20 along with their associated top words. The topic
with a -1 identifier is intended to be ignored as it represents outliers
in the data with minimal calculated importance such as ``the,'' ``of,''
and ``to.'' The top words for each topic are listed from left to right
in order of importance with the leftmost word being most representative
of the topic. This figure tells us some interesting things about the
differences in topics generated by NMF compared to BERTopic. For
example, the generated topics from BERTopic present indications of
redundancy with some topics touching on similar concepts. This can be
seen in topics 2, 7, 12, and 16 which all refer to some form of
budgeting or funding mechanism. To investigate this further, we evaluate
the magnitude of this observation by examining the intertopic distance
of each topic within our BERTopic model.

Figure 5 shows the Intertopic Distance Map for our trained BERTopic
model. Specifically, it uses multidimensional scaling to project the
topics onto a 2-dimensional plane and sizes them by term-specific
frequencies across the corpus {[}13{]}. Topics that are closer together
are similar in nature and those which are larger have more prevalence.
Reviewing this visualization confirms the observation that many of the
derived topics from BERTopic overlap one another. It also tells us that
this is consistent for almost all of the most prevalent topics in our
model when considering for size. This does not inherently mean that the
model is performing poorly, however, it will be an important
consideration when evaluating usefulness of topics towards our
application.

\subsection{Evaluation}\label{d.-evaluation}

\noindent
\begin{wrapfigure}{r}{0.45\textwidth}
  \centering
  \includegraphics[width=0.43\textwidth]{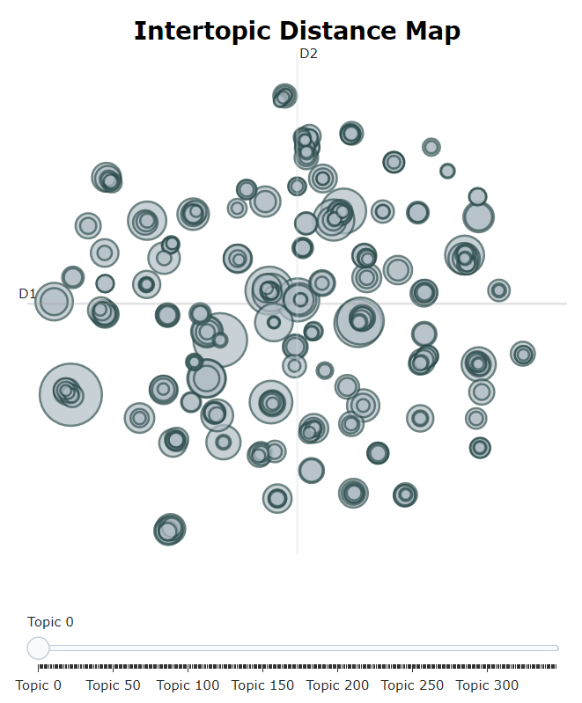}
  \caption{Intertopic Distance Map}
  \label{fig:bertopic-distance}
\end{wrapfigure}The results of our modeling are evaluated using a multi-step process to
derive the quality of the topics generated and the applicability of each
technique towards our use case. We begin this process by analyzing the
generated topics from each technique in comparison to a source of truth
data. In this case we will treat the Vision Elements from the Department
of Energy Office of Legacy Management 2020-2025 Strategic Plan {[}14{]}
as our truth data in evaluating performance. For each Vision Element, we
parse the outputs from BERTopic and NMF and document the
machine-generated topics which share resemblance to Vision Elements in
the published strategic plan. Since correlations between topics and
Vision Elements can vary in relevance, with each alignment that is
recorded we assign a correlation strength of ``weak,'' ``medium,'' or
``strong.'' After completing this analysis for all Vision Elements, we
have generated a dataset which can be evaluated for how well the
predicted topics from these techniques represent the sentiment of Vision
Elements in the published plan.

There are three primary metrics utilized in evaluating the performance of these models against our application. The first is the number of
generated topics which correlate to each Vision Element. This will tell
us the overall viability of using GAI generally, and leading topic
modeling techniques more specifically, in generating a Vision in
complement to strategic plan development. If one or more Vision Elements
have zero or few correlated topics, it indicates that these methods may
be ineffective in generating materials similar to that of a human. Next,
we measure the number of correlated topics per technique. This will
provide a perspective on which technique performs best in generating
correlated topics within the context of our specific use case. This
metric alone should be viewed thoughtfully however as certain
techniques, such as BERTopic, generate a higher number of topics by
default allowing it more opportunities for alignment to our truth data.
For this reason, we introduce a third measure, aimed at measuring the
quality of the topics generated by each technique. To accomplish this,
we calculate the distribution of correlation strengths as a percentage
for each respective technique deployed. This tells us not only the
overall quantity of correlations for each technique but what percentage
of them are weak, medium, and strong in correlation.

\begin{figure}[htbp]
  \centering
  \includegraphics[width=\linewidth]{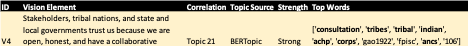}
  \caption{Example of Correlation Analysis Between Model and Vision Elements}
  \label{fig:topic-correlation}
\end{figure}

\section{Results} \label{results}

The results of this analysis reveal a total of 42 correlated topics
identified with a relatively even distribution across all six evaluated
Vision Elements. Assessing Figure 7, it can be seen that the smallest
number of correlations is 6 and the largest is 8. No Vision Elements have zero correlations, demonstrating no major gaps in performance.

\noindent
\begin{wrapfigure}{r}{0.5\textwidth}
  \centering
  \vspace{-30pt} % Optional: adjust spacing
  \includegraphics[width=0.48\textwidth]{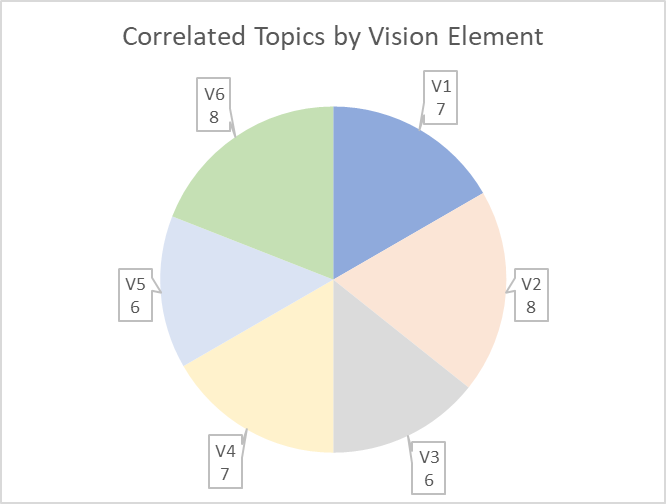}
  \caption{Correlated Topics by Vision Element}
  \label{fig:topics-by-vision}
\end{wrapfigure}
In contrast, the number of correlated topics per technique shows skewed
results. A majority of correlated topics come from the left half of
Figure 8, representing a large allocation of BERTopic topics with a weak
correlation strength. Interestingly, we see that of the two Vision
Elements that have the highest total count of correlated topics, a
majority of that count is comprised of weak correlations. No Vision
Elements demonstrate a strong correlation to NMF-generated topics and
BERTopic has an overwhelming number of more alignments than NMF,
accounting for 33 of the total of 42.

The distribution of correlation strength between techniques provides a
vantage point into the quality of generated topics. Of the total number
of correlated topics from BERTopic, roughly 24\%, demonstrate a strong
correlation to the Vision Elements whereas NMF resulted in zero strong
correlations. Both models yield roughly the same percentage of medium
strength correlations at 30\% and 33\% percent between BERTopic and NMF,
respectively. Finally, NMF showed a higher percentage of weak
correlations at 67\% versus BERTopic at 46\%. It should be noted that a
weak correlation in this context is not a bad result despite its title,
rather, it represents less correlation than other findings.

\begin{figure}[htbp]
  \centering
  \includegraphics[width=\linewidth]{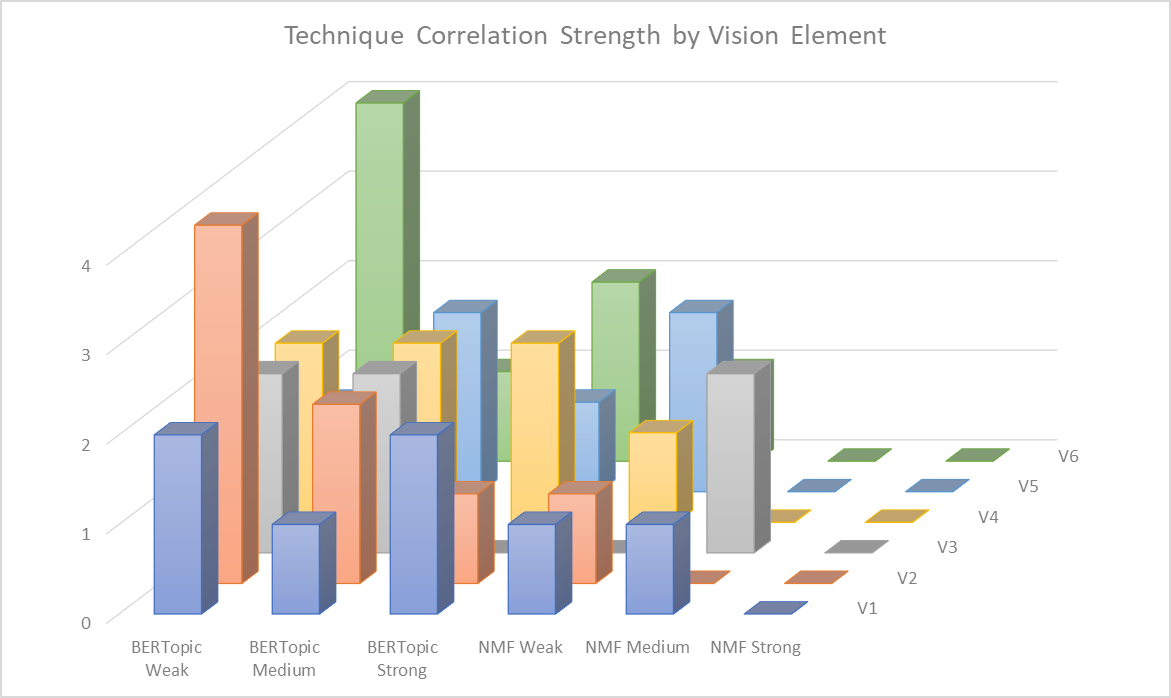}
  \caption{Technique Correlation Strength by Vision Element}
  \label{fig:correlation-strength}
\end{figure}

\section{Discussion}\label{discussion}

The results of this study prove the hypothesis that leading topic
modeling techniques can be utilized to aid in the generation of Vision
Elements in support of strategic plan development. We see evidence of
this in the number of identified correlated topics for each Vision
Element in our truth data. This tells us that the techniques deployed
are able to identify core themes represented in the 
\begin{wrapfigure}{l}{0.56\textwidth}
  \centering
  %\vspace{-5pt}  % Optional, helps adjust spacing
  \includegraphics[width=0.54\textwidth]{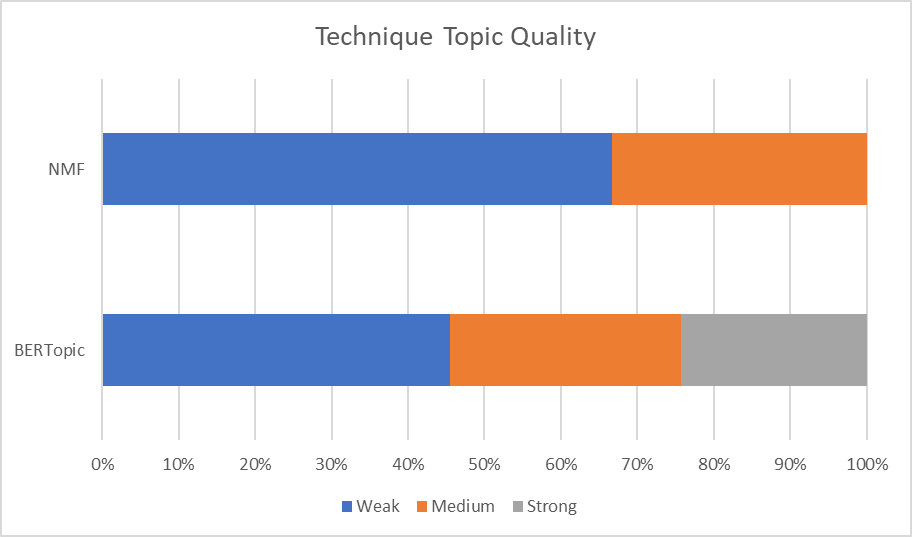}
  \caption{Technique Topic Quality}
  \label{fig:topic-quality}
\end{wrapfigure}
DOE LM Strategic
Plan. Although some of these topics were found to be weighted lower in
the list of top words for each technique, the fact that they were
identified suggests that their weighting can be adjusted to improve
their relevance to the model through additional training and
fine-tuning.

We also see that BERTopic performs better than NMF in distinguishing the
topics of relevance in the particular use case of generating Vision
Elements. This is demonstrated through the distribution of correlation
strengths between the two models. This distribution indicates a shortage
of correlated topics from NMF in comparison to BERTopic. It is
noteworthy that NMF resulted in zero strong correlations amongst all
Vision Element. In contrast, BERTopic provided full coverage of both
correlation strengths and Vision Elements, generating at least one topic
of each type for each Vision Element.

Examing the quality of correlations as an overall percentage of
correlation types only bolsters these conclusions further. More than
half of the BERTopic topics had either a medium or strong correlation to
the truth data, showing strong potential for application to our use
case. Conversely, only a third of the NMF topics reached this benchmark,
mainly due to the technique generating zero topics of strong
correlation. Through the overall number of identified correlated topics
for each Vision Element, the distribution of correlated strengths
between the two models, and the quality of correlations as a percentage
of correlation strengths, it is clear that topic modeling is a viable
approach to generating Vision Elements and that BERTopic is the highest
performing model in this application.

\section{Conclusion}\label{conclusion}

The purpose of this paper was to analyze the ability of GAI to develop a
strategic plan for large scale government organizations. We approached
this by proposing a cognitive model for how each step in the process of
developing a strategic plan can be augmented by GAI and then proving the
viability of one of the indicated modules. We concluded this analysis by
evaluating the effectiveness of leading practices in their application
to this task and identifying a leading solution with the potential to be
optimized for this purpose.

Although successful in accomplishing these objectives, this work does
not result in an immediately usable capability and more work will be
required in order to optimize the identified models and fine-tune
parameters for the intended purpose. Doing so will require additional
training data, spanning multiple government agencies. Given the rapid
development of GAI and NLP technologies, breakthroughs in new techniques
should continue to be monitored and evaluated against using the methods
outlined in this paper to determine if BERTopic continues to be the best
foundation for this application.

Future work will focus on these identified limitations in an effort to
move beyond proving theoretical viability and towards a scalable, usable
solution. Further, the other modules in the proposed cognitive model
will be evaluated for their viability in supporting the remaining areas
of strategic plan development. Achieving these milestones will produce a
holistic solution to GAI-generated strategic planning, vastly reducing
its time and cost, while enabling compliance with legal statutes such as
GPRA, for large scale government organizations.


\begin{thebibliography}{99}

\bibitem{bigelow2022}
Bigelow, S. J., \& Pratt, M. K. (2022, March 30). \emph{What is strategic planning? Definition and steps}. CIO. Retrieved from \url{https://www.techtarget.com/searchcio/definition/strategic-planning}

\bibitem{upton2001}
Upton, N., Teal, E. J., \& Felan, J. T. (2001). Strategic and business planning practices of fast growth family firms. \emph{Journal of Small Business Management}, \emph{39}(1), 60--72. \url{https://doi.org/10.1111/0447-2778.00006}

\bibitem{pmi}
Project Management Institute. (n.d.). \emph{Why good strategies fail: Lessons for the C-suite}. Retrieved June 24, 2023, from \url{https://www.pmi.org/-/media/pmi/documents/public/pdf/learning/thought-leadership/why-good-strategies-fail-report.pdf}

\bibitem{tanmy2022}
Tanmy, S., Mehul, T., \& Vineet, K. (2022). \emph{Strategy consulting market: Global opportunity analysis and industry forecast, 2021-2031}. Allied Market Research. Retrieved from \url{https://www.alliedmarketresearch.com/strategy-consulting-market-A31618}

\bibitem{schulman2023}
Schulman, J., et al. (2023). \emph{Introducing ChatGPT}. OpenAI. Retrieved from \url{https://openai.com/blog/chatgpt}

\bibitem{unhrguide}
UN HR. (n.d.). \emph{Strategic Planning Guide for Managers}. Retrieved from \url{https://hr.un.org/sites/hr.un.org/files/4.5.1.6_Strategic%20Planning%20Guide_0.pdf}

\bibitem{egger2022frontiers}
Egger, R., \& Yu, J. (2022). A topic modeling comparison between LDA, NMF, Top2Vec, and BERTopic to demystify Twitter posts. \emph{Frontiers in Sociology}, \emph{7}, 886498. \url{https://doi.org/10.3389/fsoc.2022.886498}

\bibitem{egger2022tourism}
Egger, R., \& Yu, J. (2022). Identifying hidden semantic structures in Instagram data: A topic modelling comparison. \emph{Tourism Review}, \emph{77}(4), 1234–1246. \url{https://doi.org/10.1108/TR-05-2021-0244}

\bibitem{yager2023}
Yager, K. G. (2023). Domain-specific chatbots for science using embeddings. \emph{arXiv preprint} arXiv:2306.10067

\bibitem{brown2020}
Brown, T., Mann, B., et al. (2020). Language models are few-shot learners. \emph{arXiv preprint} arXiv:2005.14165

\bibitem{egger2022book}
Egger, R. (2022). Topic modelling: Modelling hidden semantic structures in textual data. In R. Egger (Ed.), \emph{Applied Data Science in Tourism} (pp. 18). Springer. \url{https://doi.org/10.1007/978-3-030-88389-8_18}

\bibitem{grootendorst2022}
Grootendorst, M. (2022). BERTopic: Neural topic modeling with a class-based TF-IDF procedure. \emph{arXiv preprint} arXiv:2203.05794v1

\bibitem{sievert2014}
Sievert, C., \& Shirley, K. (2014). LDAvis: A method for visualizing and interpreting topics. \url{https://doi.org/10.13140/2.1.1394.3043}

\bibitem{doeplan}
U.S. Department of Energy. (2020). \emph{2020–2025 Strategic Plan}. Retrieved from \url{https://www.energy.gov/lm/articles/2020-2025-strategic-plan}

\end{thebibliography}
\end{document}